\documentclass[journal,twoside,web]{ieeecolor}
\usepackage{tmi}
\usepackage{amsmath,amssymb,amsfonts}
\usepackage{algorithmic}
\usepackage{textcomp}

\usepackage[backend=bibtex, style=ieee]{biblatex}
\usepackage{xcolor}
\usepackage{multirow}
\usepackage{graphicx}

\usepackage{hyperref}

\def\BibTeX{{\rm B\kern-.05em{\sc i\kern-.025em b}\kern-.08em
    T\kern-.1667em\lower.7ex\hbox{E}\kern-.125emX}}
\begin{document}
\title{MisMatch: Calibrated Segmentation via Consistency on Differential Morphological Feature Perturbations with Limited Labels}
\author{Mou-Cheng Xu, Yukun Zhou, Chen Jin, Marius de Groot, \\ Daniel C.~Alexander, Neil P.~Oxtoby and Joseph Jacob
\thanks{MCX is supported by GSK (BIDS3000034123) and UCL Dean's Prize. NPO is supported by a UKRI Future Leaders Fellowship (MR/S03546X/1). DCA is supported by UK EPSRC grants M020533, R006032, R014019, V034537, Wellcome Trust UNS113739. JJ is supported by Wellcome Trust Clinical Research Career Development Fellowship 209,553/Z/17/Z. NPO, DCA, and JJ are supported by the NIHR UCLH Biomedical Research Centre, UK. MCX, YKZ, CJ, NPO, DCA, JJ are with Centre for Medical Image Computing, UCL, UK. MDG is with GSK, Stevenage, UK. CJ has now moved to AstraZeneca, Cambridge, UK. Contacts: xumoucheng28@gmail.com}

}
\maketitle

\begin{abstract}
Semi-supervised learning (SSL) is a promising machine learning paradigm to address the ubiquitous issue of label scarcity in medical imaging. The state-of-the-art SSL methods in image classification utilise consistency regularisation to learn unlabelled predictions which are invariant to input level perturbations. However, image level perturbations violate the cluster assumption in the setting of segmentation. Moreover, existing image level perturbations are hand-crafted which could be sub-optimal. In this paper, we propose MisMatch, a semi-supervised segmentation framework based on the consistency between paired predictions which are derived from two differently learnt morphological feature perturbations. MisMatch consists of an encoder and two decoders. One decoder learns positive attention for foreground on unlabelled data thereby generating dilated features of foreground. The other decoder learns negative attention for foreground on the same unlabelled data thereby generating eroded features of foreground. We normalise the paired predictions of the decoders, along the batch dimension. A consistency regularisation is then applied between the normalised paired predictions of the decoders. We evaluate MisMatch on four different tasks. Firstly, we develop a 2D U-net based MisMatch framework and perform extensive cross-validation on a CT-based pulmonary vessel segmentation task and show that MisMatch statistically outperforms state-of-the-art semi-supervised methods. Secondly, we show that 2D MisMatch outperforms state-of-the-art methods on an MRI-based brain tumour segmentation task. We then further confirm that 3D V-net based MisMatch outperforms its 3D counterpart based on consistency regularisation with input level perturbations, on two different tasks including, left atrium segmentation from 3D CT images and whole brain tumour segmentation from 3D MRI images. Lastly, we find that the performance improvement of MisMatch over the baseline might originate from its better calibration. This also implies that our proposed AI system makes safer decisions than the previous methods.
\end{abstract}


\section{Introduction}
\label{section:intro}
Training of deep learning models requires a large amount of labelled data. However, in applications such as in medical image analysis, anatomic/pathologic labels are prohibitively expensive and time-consuming to obtain, with the result that label scarcity is almost inevitable. Advances in the medical image analysis field requires the development of label efficient deep learning methods and accordingly, semi-supervised learning (SSL) has become a major research interest within the community. Among the myriad SSL methods used, consistency regularisation based methods have achieved the state-of-the art in classification \cite{meanteacher,fixmatch2020,remixmatch,ssl_swa}, thus we focus on this genre in this paper. 

\begin{figure}[!ht]
\centering
\includegraphics[width=\linewidth]{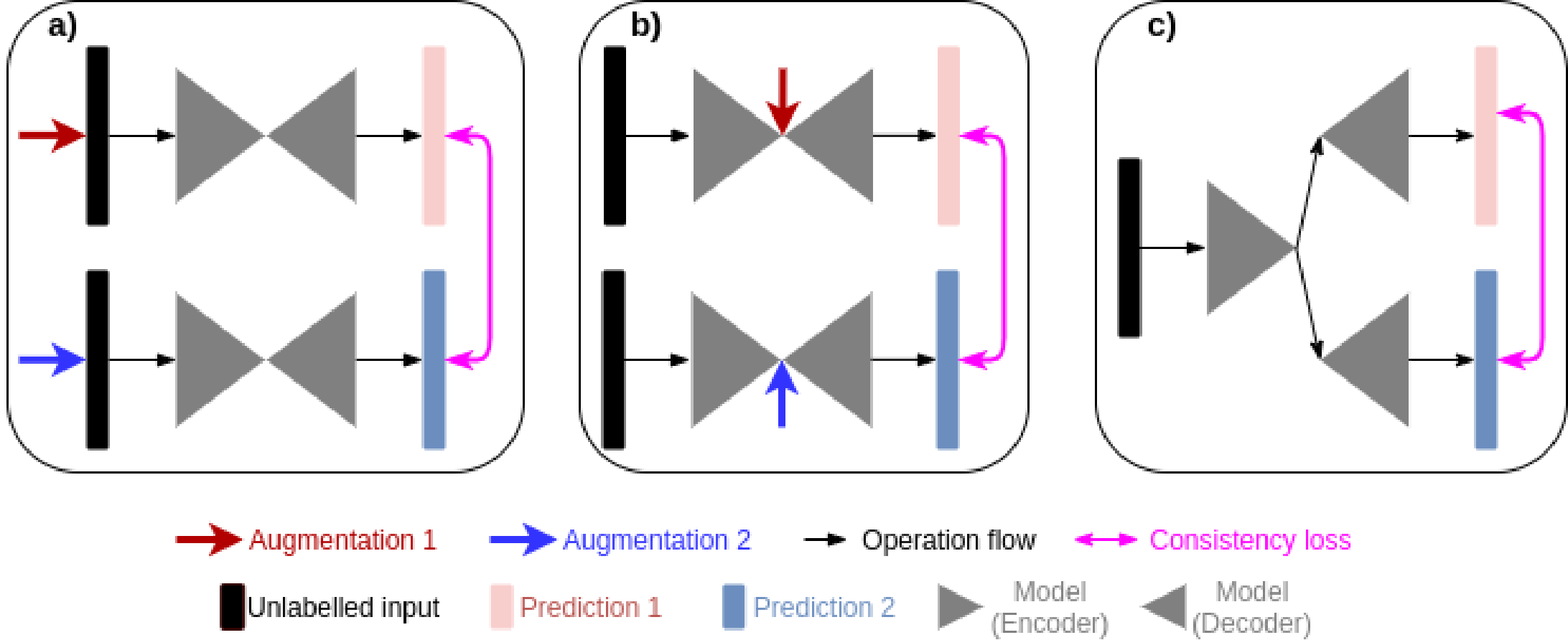}
\caption{Different strategies for consistency regularisation. (a) Previous methods \cite{meanteacher,fixmatch2020,bmvc_ssl_2020} use hand-crafted augmentation at input level to create predictions with different confidences. (b) Previous method \cite{cct_cvpr2020} uses hand-crafted augmentation at feature level to create predictions with different confidences. (c) Our method end-to-end learns to create predictions with different confidences.}
\label{fig:consistency_ssl}
\end{figure}

Existing consistency regularisation methods  \cite{meanteacher,fixmatch2020,remixmatch,ssl_swa,cct_cvpr2020,eccv_gct_2020,bmvc_ssl_2020,bmvc_ssl_2018} are mainly focusing on producing predictions which are invariant against different input level perturbations. In other words, we can interpret that consistency regularisation methods aim at training networks which generate confidence invariant predictions. For example, if we apply weak augmentation such as flipping on an input image, the model will assign a high probability of this image belonging to its correct label, hence, the prediction of the weakly augmented image is with high confidence; if we apply strong augmentation such as rotation on an input image, then the testing is much more difficult and the model might assign a low probability of this image to its correct label, therefore, such a prediction of a strongly augmented image is with low confidence. A consistency regularisation is enforced to align the paired predictions. The relationship between consistency regularisation and confidence invariant predictions imply that such networks should be having better calibration, which will be empircally verified in section~\ref{section:calibration}. However, data augmentation techniques used in existing semi-supervised learning are typically hand-crafted which might be sub-optimal. Practically, such augmentation techniques are not adaptive across pixels which may be problematic as spatial correlations amongst pixels are crucial for segmentation, e.g. neighbouring pixels might belong to the same category. Most importantly, direct adaption of input level perturbations in segmentation violates the cluster assumption which is the foundation of semi supervised learning, we will explain this issue further in later section~\ref{section:motivation}.

In this paper, we propose an end-to-end learning framework to generate predictions with different confidences. In order to change prediction confidences at a pixel-wise level in a realistic way, we use two different attention mechanisms to respectively dilate and erode foreground features which correspond to the areas of ``ground truth". A preliminary version of this manuscript has been presented at MIDL 2022 \cite{midl2022_xu}. Comparing to the previous MIDL version, we now included extra experiments on two 3D data sets using a different base network; a more detailed explanation of the motivation; a more principled method section under the guidance of the theory of effective receptive field. The code is here: \textbf{\textcolor{red}{\url{https://github.com/moucheng2017/MisMatchSSL}}}. 
Our contributions are summarised as:

\begin{itemize}
  \item We provide an intuition of the relationship between consistency regularisation and semi-supervised learning, and why consistency regularisation with data augmentation wouldn't work well in segmentation.
  \item We propose a framework called MisMatch for semi supervised segmentation, by combining differential morphological feature perturbations with consistency regularisation.
  \item We discovered that our consistency regularisation improves model calibration, leading to safer AI deployment for medicine.
  \item We intensively evaluated our framework on four medical applications including: 1) 2D segmentation of lung vessel of CT images; 2) 2D segmentation of brain tumour of MR images; 3) 3D segmentation of left atrium of MR images; 4) 3D segmentation of whole tumour from MRI images. We conclude that our consistency regularisation on feature perturbations is more effective than consistency on input level perturbations.
\end{itemize}

\section{Related work}
Popular classes of common SSL methods have been compared on a benchmark in \cite{realistic_evaluation_semi}. A direct application of smoothness assumption is called label propagation which propagate the labels to unlabelled data according to the similarity between labelled and unlabelled data \cite{label_propagation}, obviously, those similarity graphs need computationally heavy Laplacian matrices which encounter scalability issue. Another common method is called entorpy minimisation method which drives models to attain low entropy predictions on unlabelled data \cite{EntropyMinimisation} \cite{PseudoLabel}. One drawback of entropy minimisation method is the risk at overfitting leading to wrong decision boundary for data points close to low density regions (see Appendix E in \cite{realistic_evaluation_semi}). Other attempts include generative models such as the one in \cite{ssl_generative} which combines GAN in training, suffering from unstable training. The state-of-the-art methods are dominated by consistency regularisation methods becuase they are easy to use and effective across different tasks. Of the consistency regularisation methods, Mean-Teacher \cite{meanteacher} is the most representative example, containing two identical models which are fed with inputs augmented with different Gaussian noises. The first model learns to match the target output of the second model, while the second model uses an exponentially moving average of parameters of the first model. One of the state-of-the-art SSL methods \cite{remixmatch} \cite{fixmatch2020} combines entropy minimisation and consistency regularisation.


\textbf{SSL in segmentation} In semi-supervised image segmentation, consistency regularisation is commonly used \cite{miccai_roi_consistency_2019} \cite{miccai_dual_teacher_2020} \cite{IPMI_meanteacher_2019} \cite{local_global_consistency_miccai_2020} \cite{dmnet_2020} \cite{bmvc_ssl_2020} where different data augmentation techniques are applied at the input level. Another related work \cite{bmvc_ssl_2018} forces the model to learn rotation invariant predictions. Apart from augmentation at the input level, recently, feature level augmentation has gained popularity for consistency based SSL segmentation \cite{cct_cvpr2020, eccv_gct_2020}. There also have been attempts of creating perturbations via using dual network branch \cite{luo_aaai_dual_task} \cite{shi_tmi_2022}. Different from \cite{luo_aaai_dual_task} and \cite{shi_tmi_2022}, the perturbations we use are also learnt via network itself. Apart from consistency regularisation methods in medical imaging, there also have been other attempts, including the use of generative models for creating pseudo data points for training \cite{task_driven_semi_2019} \cite{miccai_data_augmentation_2020} and different auxiliary tasks as regularisation \cite{curriculum_semi_2019} \cite{multitask_semi_2019}. Since our method is a new consistency regularisation method, we focus on comparing with state-of-the-art consistency regularisation methods.

\section{Motivations}
\label{section:motivation}

\begin{figure*}[!ht]
\centering
\includegraphics[width=\linewidth]{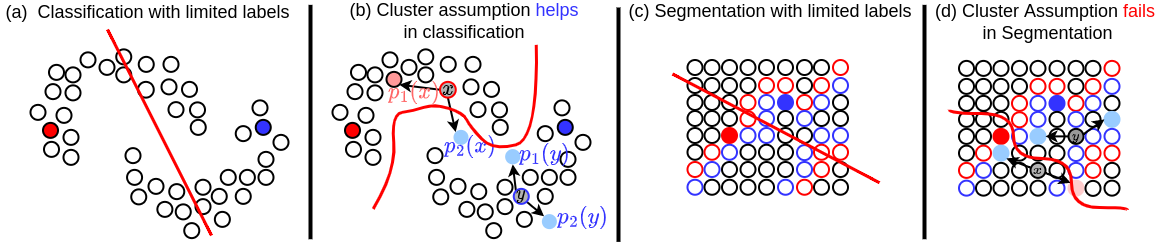}
\caption{Cluster assumptions in semi supervised classification and semi supervised segmentation. (a) In classification, limited labels will cause wrong decision boundary (red straight line), where each dot is an image. (b) In classification, cluster assumption with consistency regularisation on input level perturbations at images helps to find a better decision boundary, because low density regions of images align well with the correct decision boundary. (c) In segmentation, limited labels will cause wrong decision boundary (red straight line), where each dot is a pixel. (d) In segmentation, cluster assumption with consistency regularisation on input level perturbations at pixels will not help to find a better decision boundary, because low density regions of pixels do not align with the correct decision boundary (tight boundaries between objects).}
\label{fig:motivation}
\end{figure*}

\textbf{Cluster assumption} In this section, we will explain the cluster assumption for semi-supervised classification and how it is violated if we straightforwardly transfer existing consistency regularisation methods from classification to segmentation. The cluster assumption is a variant of smoothness assumption. The smoothness assumption states that if two data points ($x_1$ and $x_2$) are adjacent to each other, their outputs or labels ($y_1$ and $y_2$) should also be close to each other. The cluster assumption directly derives from the smoothness assumption, for example, if there is a dense population of data points in a space, then highly likely that cluster of those densely neighbouring data points are in the same class. In other words, the cluster assumptions implies there exists low density regions among different classes or different clusters of data points and the correct decision boundary should lie at the low-density regions. Equivalently, the key is to find the low-density regions which leads to rightful decision boundary. 

\textbf{Consistency with Data Augmentation in Classification} We start with a classical two moon example to explain how consistency regularisation with data augmentation works in semi-supervised classification. Each moon represents a class and each dot represents an image for semi-supervised classification. As shown in the two moons example in Fig.~\ref{fig:motivation}(a), if there are very limited labelled data points such as two data points, any decision boundary between the two labelled data points is possible, for example, the examplar decision boundary shown in Fig.~\ref{fig:motivation}(a) can wrongly classify half of the data points. The two moon example in Fig.~\ref{fig:motivation}(a) and (b) is also a perfect example for cluster assumption that the low density region between the two moons can separate the two moons from each other. In Fig.~\ref{fig:motivation}(b), let's focus on the two images $x$ and $y$ which are from upper moon class and lower moon class respectively. If we apply two random augmentations (directional arrows in Fig.~\ref{fig:motivation}(b)) on the images, we will get $p_1(x)$ and $p_2(x)$ from $x$, $p_1(y)$ and $p_2(y)$ from $y$. Since $x$ is closer to the low-density region, the augmented $x$ could across the decision boundary thereby $p_2(x)$ could be wrongly classified as the lower moon class, meanwhile, $p_1(x)$ still stays in the cluster of upper moon class. In this case, $p_1(x) != p_2(x)$ although they are derived from the same data point $x$. The difference between $p_1(x)$ and $p_2(x)$ will be more than 0 which can be back-propagated to optimise the model parameters. On the contrary, the image $y$ is closer to the centre of the cluster of lower moon class, that $p_1(y)$ and $p_2(y)$ are the same, resulting in 0 differences which does not affect the model parameters. Hence, it is easy to tell that the consistency regularisation with data augmentation makes the model parameters sensitive to the images closer to the low-density regions. This property will naturally drive the model to locate the low-density regions which happen to be the correct decision boundary.

\textbf{Consistency with Data Augmentation in Segmentation} However, consistency regularisation with data augmentation will have clear limitations in segmentation. In segmentation, as shown in Fig.~\ref{fig:motivation}(c), now we have each dot as a pixel and all of the pixels are densely distributed across the image space. In Fig.~\ref{fig:motivation}(c) and (d), we highlight the object boundary with continuous red and blue dots along the two sides of the boundary respectively. As there are hardly low-density regions between objects, it becomes hard to align the objects boundaries with low-density regions. If we have only two labelled pixels from each class, we will not be able to locate the correct decision boundary as illustrated in Fig.~\ref{fig:motivation}(c). If we apply two different augmentations on $x$ and $y$ with consistency regularisation as shown in Fig.~\ref{fig:motivation}(d), although the model can still locate the pixels which are sensitive to the consistency regularisation, due to the lack of clear low-density regions, the model will not correctly locate the right decision boundaries.

\textbf{Practical Limitations of Strong Data Augmentations in Segmentation} Common strong data augmentation techniques typically distort the spatial characterisation of the objects such as shearing. 
\begin{figure}[!htb]
    \centering
    \includegraphics[width=0.48\textwidth]{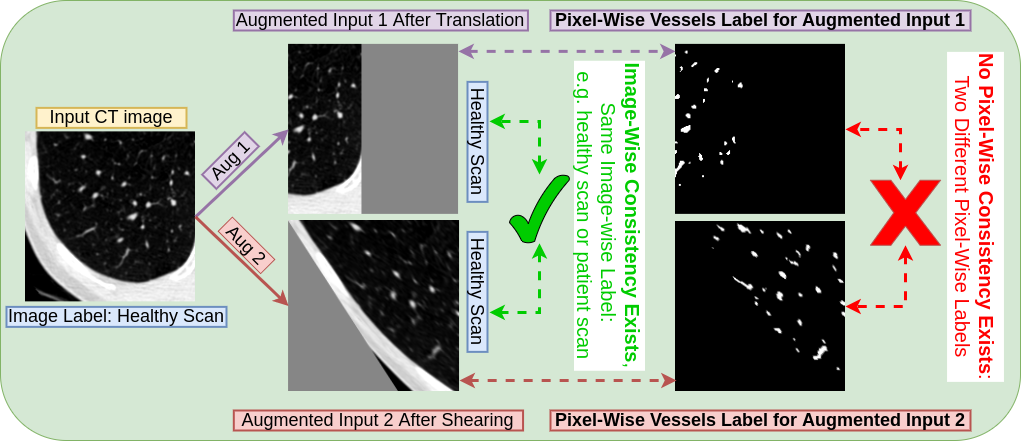}
    \caption{Strong data augmentations (e.g. shearing) change  pixel-wise labels therefore they might make pixel-wise consistency regularisation not feasible for segmentation.}
    \label{fig:prob_augmentation}
\end{figure}
As shown in Fig.\ref{fig:prob_augmentation}, the image-wise label stay the same, regardless of the data augmentation is applied. However, strong data augmentation will modify the pixel-wise labels, leading to difficulty of applying consistency regularisation at pixel-wise if two different strong data augmentations are applied on the same image. To avoid this practical issue, specific strong data augmentation such as CutMix was chosen in order to use consistency regularisation in segmentation \cite{bmvc_ssl_2020}. In our paper, we propose an alternative solution. We use augmentation at the feature level in lieu of augmentation of the data level, to completely avoid this practical issue.

\textbf{Proposal} Although the low-density regions do not align with the objects boundaries anymore, a few evidences in \cite{cct_cvpr2020, bmvc_ssl_2020} suggested that the low-density regions actually align well with the objects boundaries in the feature space. Meaning that it might be possible to use consistency regularisation on the predictions which are invariant to feature perturbations to identify the correct decision boundaries in segmentation. This directly inspired us to focus on feature perturbations in our work that we want to design learnable feature perturbations which are realistic and semantically meaningful. More specifically, we decide to apply morphological-alike perturbations on the features. In the following sections, we show how to use inductive biases of neural network topology to ask the networks to end-to-end learn morphological feature perturbations.  

\section{Methods}
\label{section:method}
\subsection{Background: ERF and the foreground} 
\textbf{Effective Receptive Field} We introduce how to control the size of the foreground features by controlling the effective receptive field (ERF). ERF \cite{erf} measures the size of the effective area at the centre of receptive field and it impacts the most on the prediction confidence of the central pixel of the receptive field, which should overlap with the foreground objects with the highest confidence at the foreground central pixel. If we want to apply morphological operations on features of foreground objects, equivalently, we need to adjust the ERF on the foreground. As found in \cite{erf}, larger ERF means the model can effectively take a larger area of the image into account during inference of decision making, resulting in higher prediction confidence at the centre, meanwhile, smaller ERF leads to less confident prediction on the central pixel due to the lack of visual information of neighbouring pixels. More importantly, ERF is highly affected by the network architecture. In particular, the dilated convolutional layer can increase the ERF to an extent dependent on the dilation rate \cite{erf}. Skip-connections conversely can shrink the ERF, though the extent of this effect is as yet unknown \cite{erf}. We are therefore inspired by \cite{erf} to design a network to control the ERF, in order to deliberately change the prediction confidence to morph the foreground features.

\begin{figure*}[!ht]
\centering
\includegraphics[width=\textwidth]{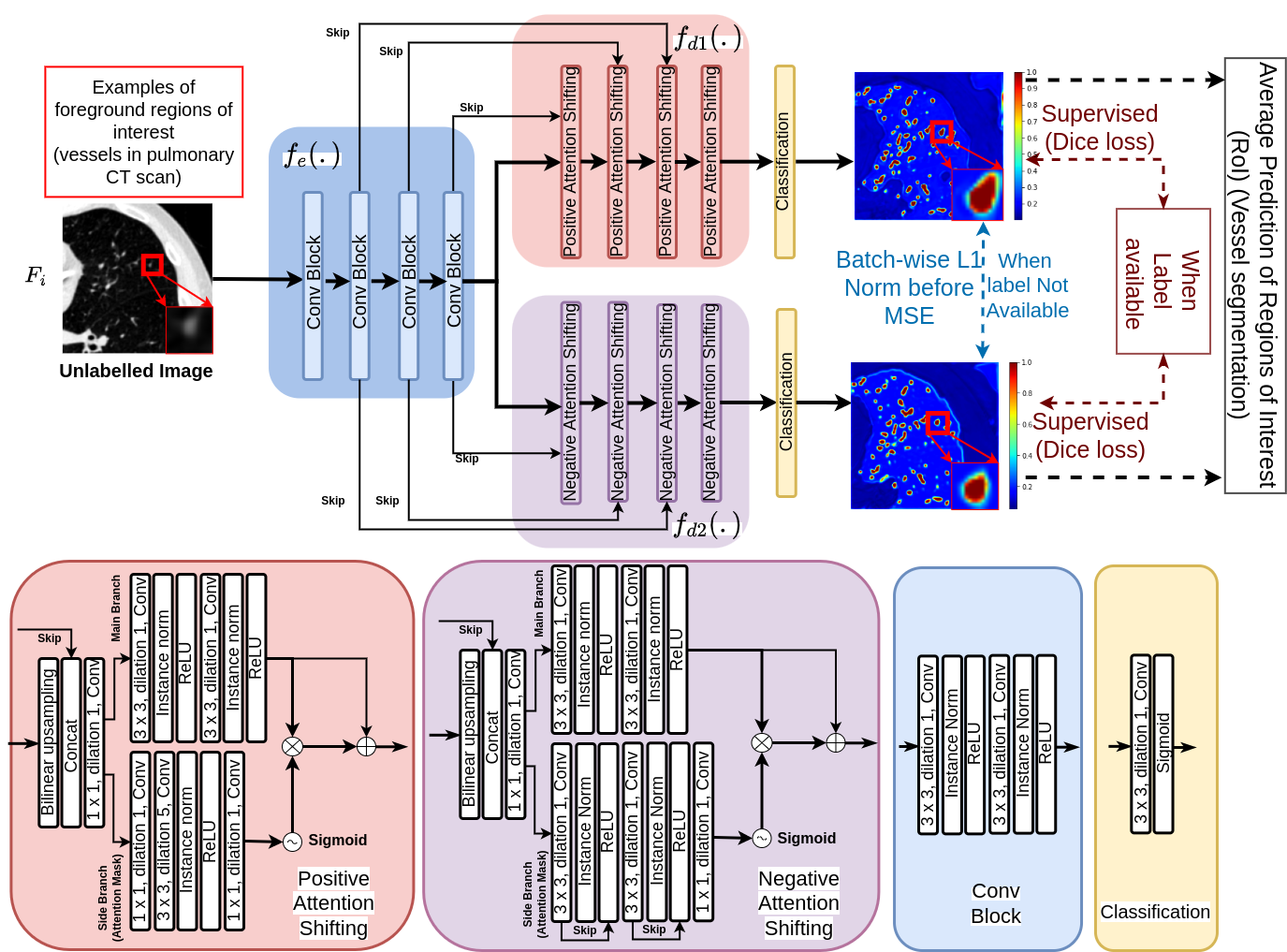}
\caption{MisMatch (U-net based): decoder $f_{d1}$ leads to dilated high confidence detection of foreground and decoder $f_{d2}$ leads to eroded high confidence detection of foreground. The final prediction is the average between outputs of $f_{d1}$ and $f_{d2}$. Any other encoder-decoder segmentation network could be used.}
\label{fig:mismatch}
\end{figure*}

\textbf{Overview of MisMatch} In this paper, we learn to realistically morph the foreground features by controlling the ERF for consistency regularisation. In order to create a paired predictions with different confidences for consistency regularisation, our strategy is to dilate the foreground features and erode the foreground features, we also compare our strategy with other possible strategies in an ablation study in later section \ref{section:results}. As introduced in the last section, the prediction confidence can be affected by the ERF while the ERF is decided by the network topology. More specifically, we use the dilated convolutional layer to raise the ERF on one hand to dilate the foreground features, and we use skip-connections to decrease the ERF on the other hand to erode the features of foreground. However, we do not know how much confidence should be changed at each pixel. To address this, we introduce soft attention mechanism to learn the magnitude of the confidence change for each pixel. Now we introduce how we achieve this in the next section. 

\textbf{Differences between proposed methods and classical morphological operations} We also would like to highlight the difference between our approach at feature space and the classical morphological operations at image space. Traditional morphological operations simply remove/add boundary pixels using local neighbouring information which is not differentiable, in contrast, our approach is differentiable and can be fully integrated in neural networks.

\subsection{Architecture of Mismatch}
As shown in Fig.\ref{fig:mismatch}, MisMatch is a framework which can be integrated into any encoder-decoder based segmentation architecture. In this section, we use 2D U-net \cite{unet} due to its popularity in medical imaging, although later we also have an experiment using a MisMatch based on a 3D V-net. Our U-net based MisMatch (\textbf{Fig \ref{fig:mismatch}}) has two components, an encoder ($f_e$) and a two-head decoder ($f_{d1}$ and $f_{d2}$). The first decoder ($f_{d1}$) comprises of a series of \textit{Positive Attention Shifting Blocks}, which shifts more attention towards the foreground area, resulting in dilating high-confidence predictions on the foreground. The second decoder ($f_{d2}$) containing a series of \textit{Negative Attention Shifting Blocks}, shifts less attention towards the foreground, resulting in eroding high-confidence predictions on the foreground. 

\begin{figure*}[!ht]
\centering
\includegraphics[width=\textwidth]{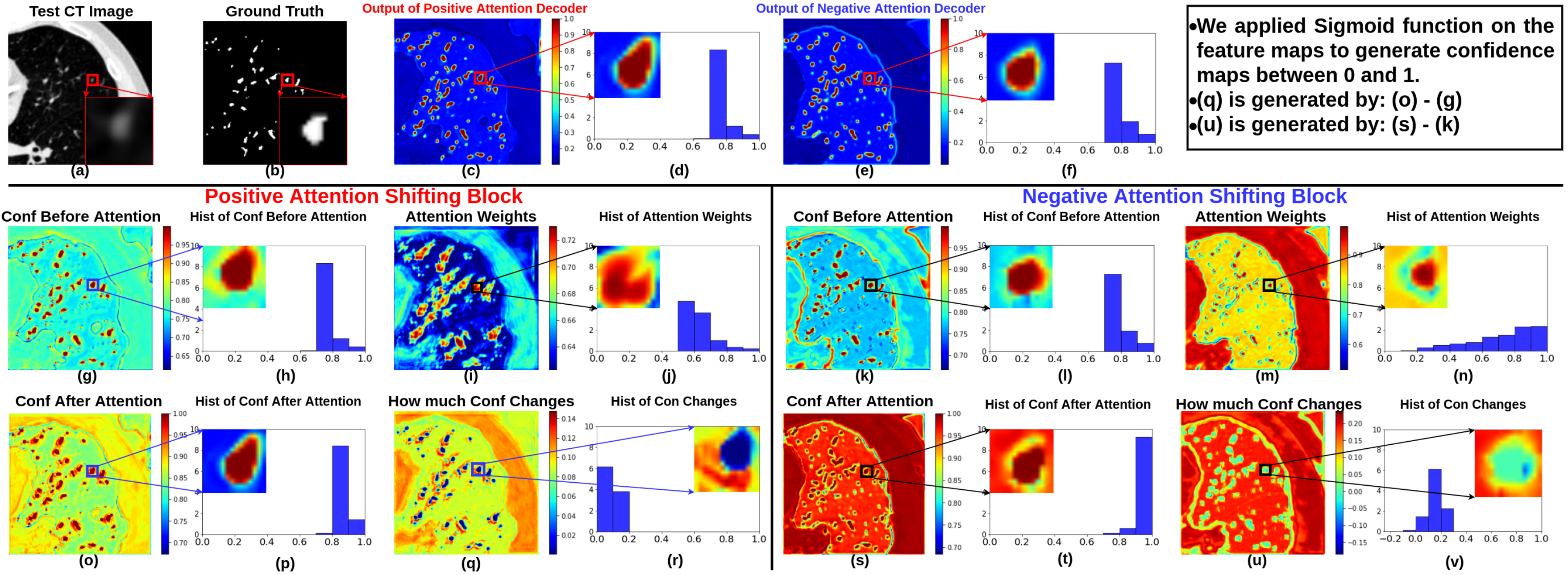}
\caption{Visulisation of confidences in the last positive attention shifting decoder and the last negative attention shifting decoder. We focus on the zoomed-in regions on the foreground area containing one vessel. As shown in (p) from the positive attention shifting block, the confidence on the foreground has been raised that even the surrounding areas outside the foreground contour have a high confidence as the foreground. Meanwhile, the confidence on the centre of the foreground remains high as the confidence is already high at the central areas before positive attention is applied. As for the negative attention shifting block, as shown in (t), the confidence on the peripheral areas on the foreground has been decreased as shown in green and blue colours. Additionally, the difference between before and after negative attention in (v) further confirms the effectiveness of the negative attention, as the difference values are negative (see the colour bars in (v)). As shown in the attention weights in (j) and (n), both the attention blocks focus on changing the confidence on the edges of the foreground, this is because the edges are normally the most ambiguous areas.}
\label{fig:attention_visulisation_mismatch}
\end{figure*}

\subsection{Positive Attention Shifting Block}
\label{pasb}
Positive Attention Shifting Block aims at increasing the ERF of the foreground, therefore dilating the foreground features. In a standard U-net, a block ($f(.)$) in the decoder comprises two consecutive convolutional layers with kernel size ($K$) 3 followed by ReLU and normalisation layers. If the input of $f(.)$ is $x$ and the output of $f(.)$ is $f(x)$, to increase the ERF of $f(x)$, we would aim to generate an attention mask with a larger ERF than the ERF of $f(x)$. To do so, we add a parallel side branch $f'(.)$ next to the main branch $f(.)$. The side branch intakes $x$ but outputs $f'(x)$ with a larger ERF. We apply Sigmoid on the output of the side branch as an attention mask to increase the confidence of $f(x)$. The new block containing both $f(.)$ and $f'(.)$ is our proposed Positive Attention Shifting Block (PASB). The side branch of the PASB is a dilated convolutional layer with dilation rate 5.

\subsubsection{ERF size in Positive Attention Shifting Block}
Given the size of ERF of $n^{th}$ layer as, $\sqrt{n}$ \cite{erf}, which is the input $x$, as output from the previous layer. The ERF of $f(x)$ is $ERF_{f(x)} = K\sqrt{n+2}$. To make sure the ERF of $f'(x)$ is larger than $K\sqrt{n+2}$:
\begin{equation}
\label{proof1}
\begin{split}
&\frac{ERF_{f'(x)}}{ERF_{f(x)}} = \frac{K'}{K} \sqrt{\frac{1}{1 + \frac{1}{n+1}}} > \lim_{n \to + 0} \frac{K'}{K} \sqrt{0.5} > 1
\end{split}
\end{equation}
From Eq\ref{proof1}, we find $K' > \frac{1}{\sqrt{0.5}} K \approx 1.5 K$. We double the condition as our design choice, then $K'$ is 9 when $K = 3$. However, the large kernel sizes significantly increase model complexity. To avoid this, we use a dilated convolutional layer to achieve $K'$ at 9, which requires a dilation rate 5. As the side branch has a larger ERF than the main branch, it can raise the confidence on the foreground of the main branch. Previous work \cite{mcx_bmvc_2020, cvpr_dilated_ssl} has reported similar uses of a dilated convolutional layer to increase the ERF for other applications, without explaining the rationale for their use. See visual evidence in Fig~\ref{fig:mismatch}(q) and (r).

\subsection{Negative Attention Shifting Block}
\label{nasb}
Negative Attention Shifting Block aims at decreasing the ERF on the foreground, therefore eroding the foreground features. Following PASB, we design the Negative Attention Shifting Block (NASB) again as two parallel branches. In NASB, we aim to shrink the ERF of the $f(x)$ in order to produce a smaller ERF than the one from the main branch. In the side branch in NASB, we use the same architecture as the main branch, but with skip-connections as skip-connections restrict the growth of the ERF with increasing depth \cite{erf}. 

\subsubsection{ERF size in Negative Attention Shifting Block}
Neural networks with residual connections are equivalent to an ensemble of networks with short paths where each path follows a binomial distribution \cite{unravel}. If we define $p$ as the probability of the model going through a convolutional layer and $1 -p$ as the probability of the model skipping the layer, then each short path has a portion of $\binom{N}{k} p^k (1-p)^{n-k}$, contributing to the final ERF. If we assume $p$ is 0.5, the ERF of the side branch is guaranteed to be smaller than the ERF of the main branch, see Eq.\ref{prop2}. 
\begin{equation}
\label{prop2}
\begin{split}
&\frac{ERF_{f'(x)}}{ERF_{f(x)}} =0.25\sqrt{\frac{1}{1 + \frac{2}{n}}} + 0.5 \sqrt{\frac{1}{1 + \frac{1}{n+1}}} + 0.25 \\
& < \lim_{n \to + \infty} 0.25 + 0.5 + 0.25 =1
\end{split}
\end{equation} 
As the side branch has a smaller ERF than the main branch, it can reduce the confidence on the foreground of the main branch. See visual evidence in Fig~\ref{fig:mismatch}(u) and (v).

\subsection{Loss Functions}
\label{section:loss}
For experiments on BRATS 2018 and CARVE 2014, We use a streaming training setting to avoid over-fitting on limited labelled data so the model doesn't repeatedly see the labelled data during each epoch. When a label is available, we apply a standard Dice loss \cite{dice_loss} between the output of each decoder and the label. When a label is not available, we apply a mean squared error loss between the outputs of the two decoders. This consistency regularisation is weighted by hyper-parameter $\alpha$. For experiments on LA 2018, we train simultaneously on labelled and unlabelled images by combine consistency regularisation loss with Dice loss. 

\section{Experiments}
We perform a few sets of experiments: 1) comparisons with baselines including supervised learning and state-of-the-art SSLs \cite{fixmatch2020,meanteacher,multitask_semi_2019,cct_cvpr2020} using either data or feature augmentation; 2) investigation of the impact of the amount of labelled data and unlabelled data on MisMatch performance; 3) ablation study of the decoder architectures; 4) ablation study on the hyper-parameter such as $\alpha$ 

\subsection{Data sets \& Pre-processing} 
\label{section:datasets}
\textbf{CARVE 2014} The Classification of pulmonary arteries and veins (CARVE) dataset \cite{carve2014} has 10 fully annotated non-contrast low-dose thoracic CT scans. Each case has between 399 and 498 images, acquired at various spatial resolutions between (282 x 426) to (302 x 474). 10-fold cross-validation on the 10 labelled cases is performed. In each fold, we split cases as: 1 for labelled training data, 3 for unlabelled training data, 1 for validation and 5 for testing. We only use slices containing more than 100 foreground pixels. We prepare datasets with differing amounts of labelled slices: 5, 10, 30, 50, 100. We crop 176 $\times$ 176 patches from four corners of each slice. Full label training uses 4 training cases. Normalisation was performed at case wise. 

\textbf{BRATS 2018} BRATS 2018 \cite{brats2015} has 210 high-grade glioma and 76 low-grade glioma MRI cases, each case containing 155 slices. We focus on binary segmentation of whole tumours in high grade cases. We randomly select 1 case for labelled training, 2 cases for validation and 40 cases for testing. We centre crop slices at 176 $\times$ 176. For labelled training data, we extract the first 20 slices containing tumours with areas of more than 5 pixels. To see the impact of the amount of unlabelled training data, we use 3100, 4650 and 6200 slices respectively. Case-wise normalisation was performed and all modalities were concatenated. We train each model 3 times and take the average.

\textbf{LA 2018} Atrial Segmentation Challenge Data set \cite{xiong2020global_la} has 100 volumes of 3D gadolimium-enhanced MR scans with corresponding left atrium segmentation masks. Each scan is isotropic with resolution at 0.625 x 0.625 x 0.625 $mm^3$. We follow \cite{miccai_uamt} and split 100 scans into 80 for training and 20 for testing. We also directly use the pre-processing from \cite{miccai_uamt} to normalise the centre crop each scan.

\textbf{Task 01 Brain Tumour} Task01 Brain Tumour from Medical Segmentation Decathlon consortium \cite{med_decathlon} is based on BRATS 2017 with different naming format from BRATS 2018. Each case in The Task01 Brain Tumour has 155 slices with 240 x 240 spatial dimension. We merge all of the tumour classes into one tumour class for simplicity. We do not apply centre cropping in the pre-processing here. In the training, we randomly crop volumes on the fly with size of 96 x 96 x 96. We separate the original training cases as labelled training data and testing data. We use the original testing cases as unlabelled data. For the labelled training data, we use 8 cases with index number from 1 to 8. We have 476 cases for testing and 266 cases for unlabelled training data. We apply normalisation with statistics of intensities across the whole training data set. We keep all of the MRI modalities as 4 channel input. 


\begin{table*}[!htb]
\caption{MisMatch (MM) vs Baselines on CARVE. Metric is Intersection over Union (IoU).}
  \label{tab:carve}
  \centering
  \resizebox{\textwidth}{!}
  {
  \begin{tabular}{ c  | c   c | c  c  c  c  c  c  }
    \hline
     & \multicolumn{2}{c|}{\bfseries Supervised} & \multicolumn{6}{c}{\bfseries Semi-Supervised} \\
    \hline
    \bfseries Labelled & \bfseries Sup1 &\bfseries Sup2 & \bfseries MTA &\bfseries MT
    &\bfseries FM &\bfseries CCT &\bfseries Morph &\bfseries MM \\
    \bfseries Slices & \cite{unet}(2015) & Ours(2021) & \cite{multitask_semi_2019}(2019) & \cite{meanteacher}(2017)
    & \cite{fixmatch2020}(2020) & \cite{cct_cvpr2020}(2020) & 2021 & \textbf{Ours(2021)} \\
    \hline
    5 & 48.32$\pm$4.97 & 50.75$\pm$2.0 & 54.91$\pm$1.82 & \textcolor{blue}{56.56$\pm$2.38} & 49.30$\pm$1.81 & 52.54$\pm$1.74 & 52.93$\pm$2.19 &
    \textcolor{red}{\textbf{60.25$\pm$3.77}} \\
    10 & 53.38$\pm$2.83 & 55.55$\pm$4.42 & 57.78$\pm$3.66 & \textcolor{blue}{57.99$\pm$2.57} & 51.53$\pm$3.72 & 55.25$\pm$2.52 & 57.08$\pm$2.96 & \textcolor{red}{\textbf{60.04$\pm$3.64}} \\
    30 & 52.09$\pm$1.41 & 53.98$\pm$4.42 & \textcolor{blue}{60.78$\pm$4.63} & 60.46$\pm$3.74 & 55.16$\pm$5.93 & 60.81$\pm$4.09 & 60.19$\pm$4.97 & \textcolor{red}{\textbf{63.59$\pm$4.46}} \\
    50 & 60.69$\pm$2.51 & 64.79$\pm$3.46 & \textcolor{blue}{68.11$\pm$3.39} & 67.21$\pm$3.05 & 62.91$\pm$6.99 & 65.06$\pm$3.42 & 64.88$\pm$3.25 & \textcolor{red}{\textbf{69.39$\pm$3.74}} \\
    100 & 68.74$\pm$1.84 & \textcolor{blue}{73.1$\pm$1.51} & 72.48$\pm$1.61 & 71.48$\pm$1.57 & 72.58$\pm$1.84 & 72.07$\pm$1.75 & 72.11$\pm$1.88 & \textcolor{red}{\textbf{74.83$\pm$1.52}} \\
    \hline
    Param. (M)  & 1.8 & 2.7 & 2.1 & 1.88 & 1.88 & 1.88 & 2.54 & 2.7\\
    \hline
    Infer.Time(s) & 4.1e-3 & 1.8e-1 & 7.2e-3 & 4.3e-3 & 4.5e-3 & 1.5e-1 & 8e-3 & 1.8e-1 \\
    \hline
    \end{tabular}
    }
    
\end{table*}

\begin{table*}[!htb]
\caption{MisMatch (MM) vs Baselines on BRATS. Metric is Intersection over Union (IoU).}
  \label{tab:brats}
  \centering
  \resizebox{\textwidth}{!}
  {
  \begin{tabular}{ c |  c  c | c  c  c  c  c  c}
    \hline
    & \multicolumn{2}{c|}{\bfseries Supervised} & \multicolumn{6}{c}{\bfseries Semi-Supervised} \\
    \hline
    \bfseries  Unlabelled & \bfseries Sup1 &\bfseries Sup2 & \bfseries MTA &\bfseries MT
    &\bfseries FM &\bfseries CCT &\bfseries Morph &\bfseries MM \\
    \bfseries  Slices & \cite{unet}(2015) & Ours(2021) & \cite{multitask_semi_2019}(2019) & \cite{meanteacher}(2017)
    & \cite{fixmatch2020}(2020) & \cite{cct_cvpr2020}(2020) & 2021 & \textbf{Ours(2021)} \\
    \hline
     3100 & 53.74$\pm$10.19 & 55.76$\pm$11.03 & 50.53$\pm$8.76 & 55.29$\pm$10.21 & \textcolor{blue}{57.92$\pm$12.35} & 56.61$\pm$11.7 & 53.88$\pm$9.99 & \textcolor{red}{\textbf{58.94$\pm$11.41}} \\
     4650 & 53.74$\pm$10.19 & 55.76$\pm$11.03 & 47.36$\pm$6.65 & \textcolor{blue}{58.32$\pm$12.07} & 54.29$\pm$9.69 & 56.94$\pm$10.93 & 55.82$\pm$11.03 & \textcolor{red}{\textbf{60.74$\pm$12.96}} \\
     6200 & 53.74$\pm$10.19 & 55.76$\pm$11.03 & 50.11$\pm$8.00 & 56.92$\pm$12.20 & 56.78$\pm$11.39 & \textcolor{blue}{57.37$\pm$11.74} & 54.5$\pm$9.75 &  \textcolor{red}{\textbf{58.81$\pm$12.18}} \\
     \hline
    \end{tabular}
    }
\end{table*}

\subsection{Implementation}
We use Adam optimiser \cite{adam}. Hyper-parameters are: $\alpha=0.002$, batch size 1 (GPU memory: 2G), learning rate 2e-5, 50 epochs. Each complete training on CARVE takes about 3.8 hours. The final output is the average of the outputs of the two decoders. In testing, we take an average of models saved over the last 10 epochs across experiments. Our code is implemented using Pytorch 1.0 \cite{pytorch}. 

\subsection{Baselines} 
In the current study the backbone is a 2D U-net \cite{unet} with 24 channels in the first encoder. To ensure a fair comparison we use the same U-net as the backbone across all baselines. The first baseline utilises supervised training on the backbone, is trained with labelled data, augmented with flipping and Gaussian noise and is denoted as ``Sup1''. To investigate how unlabelled data improves performance, our second baseline ``Sup2'' utilises supervised training on MisMatch, with the same augmentation. Because MisMatch uses consistency regularisation, we focus on comparisons with five consistency regularisation SSLs: 1) ``mean-teacher'' (MT) \cite{meanteacher}, with Gaussian noise, which has inspired most of the current state-of-the-art SSL methods; 2) the current state-of-the-art model called ``FixMatch'' (FM) \cite{fixmatch2020}. To adapt FixMatch for a segmentation task, we use Gaussian noise as weak augmentation and ``RandomAug'' \cite{random_augmentation} without shearing for strong augmentation. We do not use shearing for augmentation because it impairs spatial correspondences of pixels of paired dense outputs; 3) a state-of-the-art model with multi-head decoder  \cite{cct_cvpr2020} for segmentation (CCT), with random feature augmentation in each decoder \cite{cct_cvpr2020}. This baseline is also similar to models recently developed \cite{bmvc_ssl_2020, eccv_gct_2020}; 4) a further recent model in medical imaging \cite{multitask_semi_2019} using image reconstruction as an extra regularisation (MTA), augmented with Gaussian noise; 5) a U-net with two standard decoders, where we respectively apply erosion and dilation on the features in each decoder, augmented with Gaussian noise (Morph)''; 6) an uncertainty aware mean-teacher based SSL segmentation model \cite{miccai_uamt}. Our MisMatch model has been trained without any augmentation.

\section{Segmentation Results}
\label{section:results}

\begin{figure*}[!htb]
    \centering
    \begin{center}
        \includegraphics[width=\textwidth]{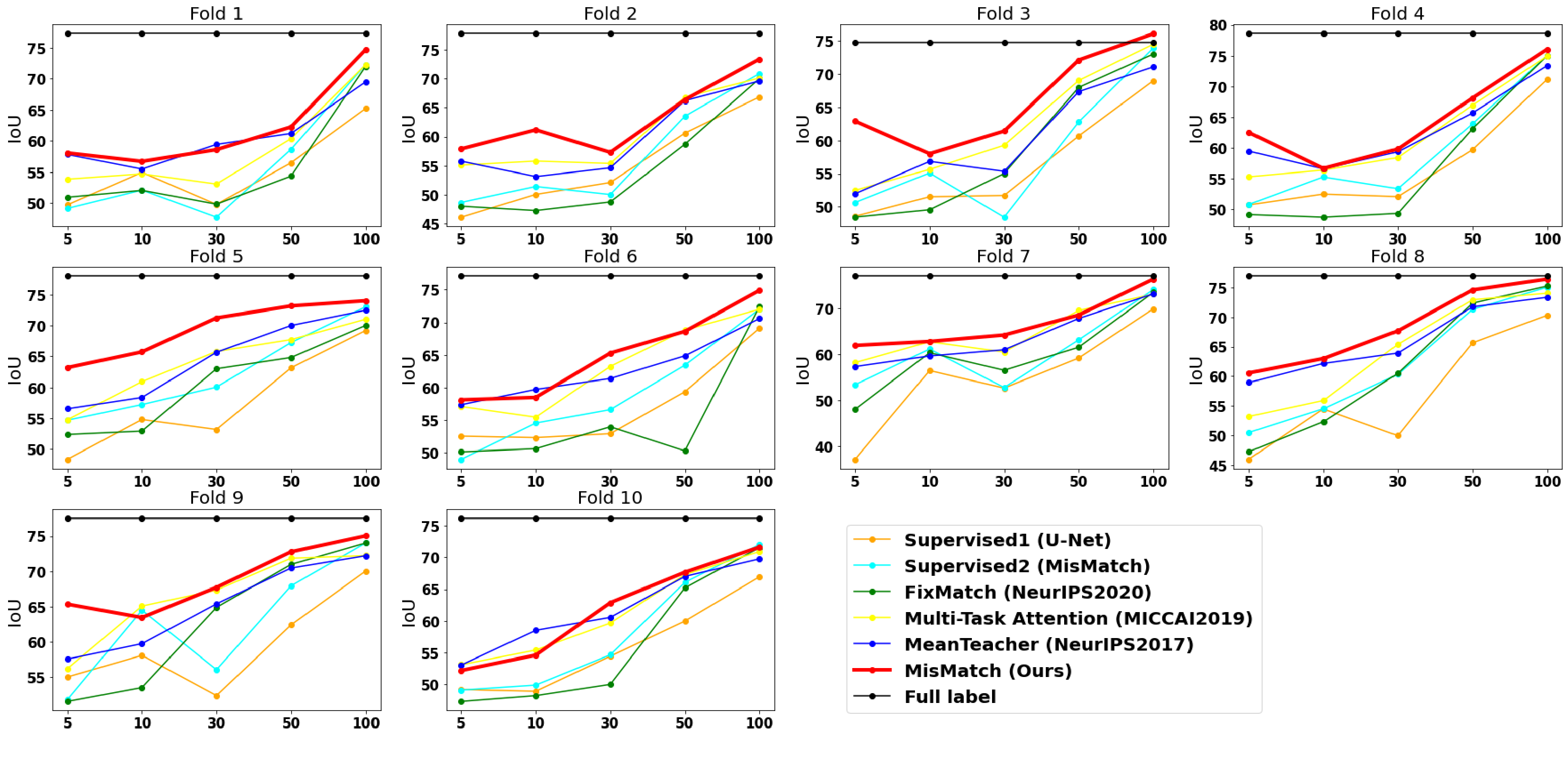}
    \end{center}
    \caption{Full results of 10 fold cross-validation on CARVE. X-axis: number of labelled slices. Y-axis: IoU}
    \label{fig:cross_validation}
\end{figure*}

\begin{figure}[!htb]
    \centering
    \begin{center}
        \includegraphics[width=0.48\textwidth]{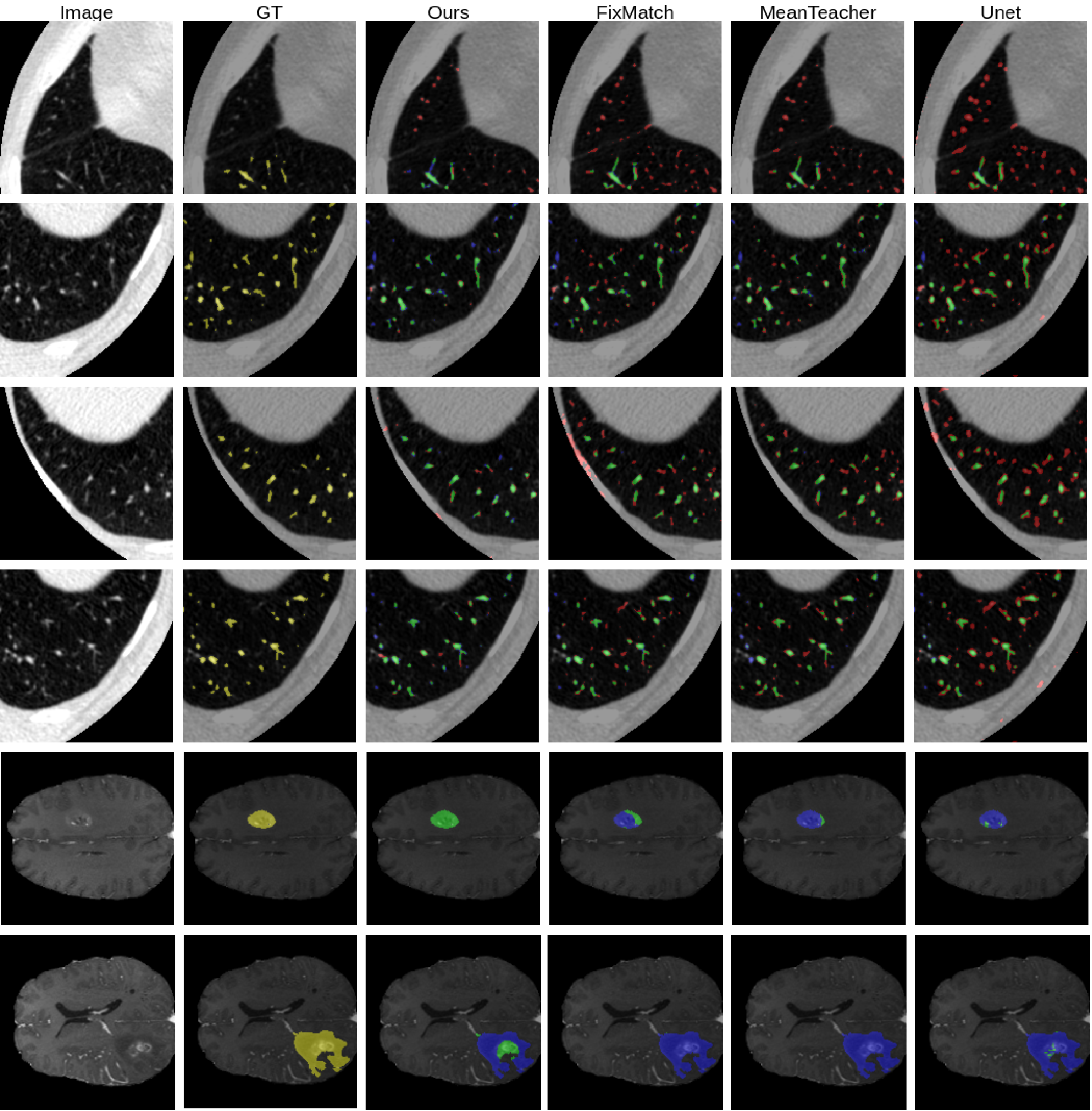}
    \end{center}
    \caption{Visual results. Yellow: ground truth. Red: False Positive. Green: True Positives. Blue: False Negatives. Row 1-4: CARVE. Row 5-6: BRATS}
    \label{fig:visual_result}
\end{figure}

MisMatch consistently and substantially outperforms supervised baselines, the improvement is especially obvious in low data regime. For example, on 5 labelled slices with CARVE, MisMatch achieves 24\% improvement over Sup1. MisMatch consistently outperforms previous SSL methods \cite{fixmatch2020,meanteacher,multitask_semi_2019,cct_cvpr2020} in Table \ref{tab:carve}, across different data sets. Particularly, there exists statistical difference between Mismatch and other baselines when 6.25\% labels (100 slices comparing to 1600 slices of full label) are used on CARVE (Table \ref{pvalue}). Qualitatively, we observed in Fig \ref{fig:visual_result} that, the main performance boost of MisMatch comes from the reduction of false positive detection and the increase of true positive detection. 

Interestingly, we found that Sup2 (supervised training on MisMatch without unlabelled data) is a very competitive baseline comparing to previous semi-supervised methods. This might imply that MisMatch can potentially help with the supervised learning as well. 

We also found data diversity of training data highly affects the testing performance (Fig \ref{fig:cross_validation}) in cross-validation experiments. For example, in fold 3, 7 and 8 on CARVE, MisMatch outperforms or performs on-par with the full label training, whereas in the rest folds, MisMatch performs marginally inferior to the full label training. Additionally, more labelled training data consistently produces a higher mean IoU and lower standard deviation (Table \ref{tab:brats}). Lastly, we noticed more unlabelled training data can help with generalisation, until it dominates training and impedes performance (Table \ref{tab:brats}).

\begin{figure}[!t]
    \centering
    \begin{center}
        \includegraphics[width=0.48\textwidth]{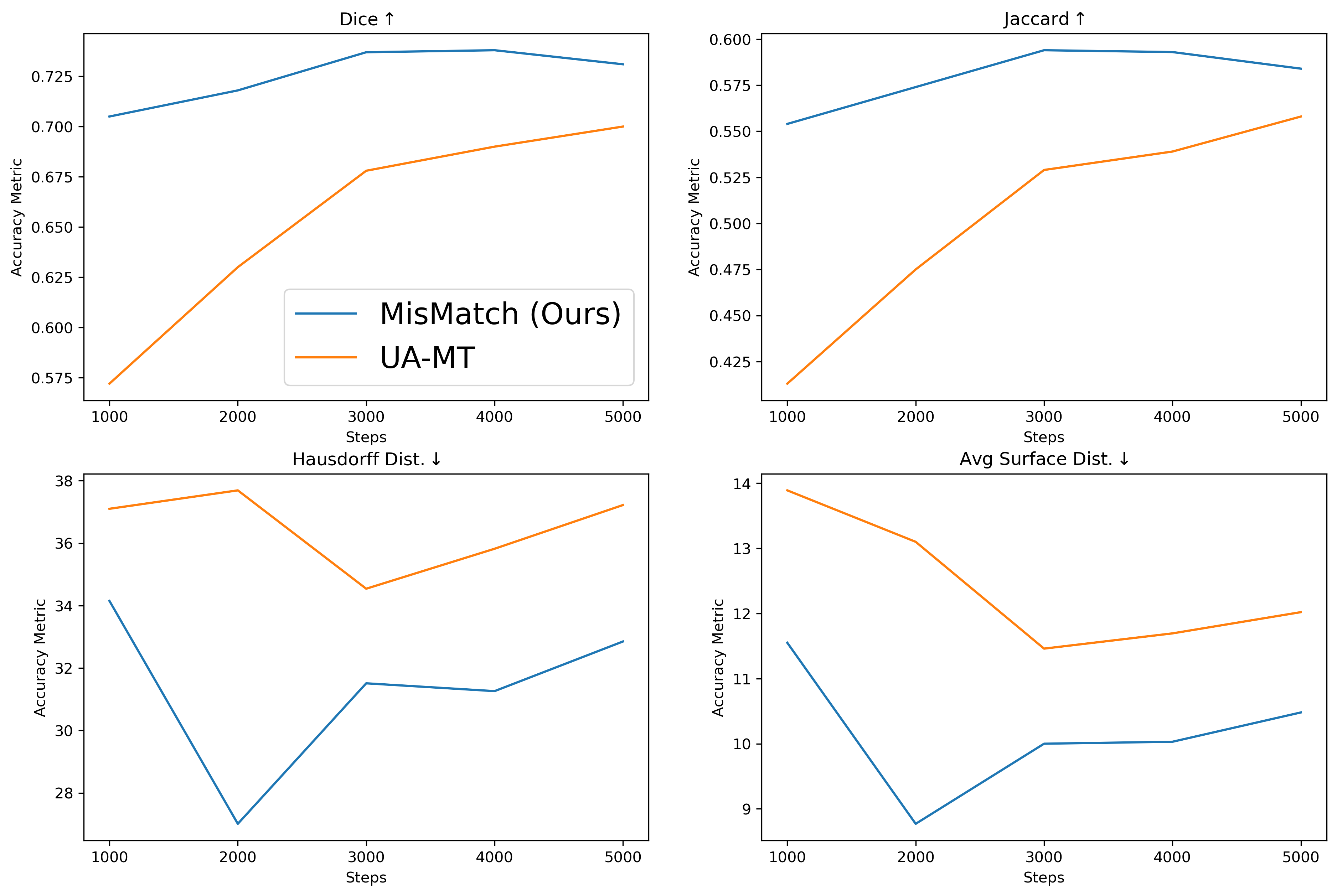}
    \end{center}
    \caption{Results on LA 2018 between UA-MT and MisMatch with 2 labelled cases, lr 0.01, batch 4, consistency 1 and network width 8. This further confirms that consistency regularisation on feature perturbations is more effective than consistency on input perturbations.}
    \label{fig:results_la}
\end{figure}

We further verify that consistency regularisation on feature perturbations is better than consistency regularisation on input perturbations by comparing MisMatch against UA-MT \cite{miccai_uamt} which is an representative example of the methods using input perturbations. We compare MisMatch against UA-MT on two 3D datasets left atrium and whole tumour areas (see section \ref{section:datasets}). On the segmentation on left atrium, our method not just outperform UA-MT but also converges faster, as illustrated in Fig.\ref{fig:results_la}. 

During testing of trained models on the whole tumour segmentation from the Task01 Brain Tumour data set\cite{med_decathlon}, we noticed one emerging property of our model that the our model achieves better performance when it is tested on volumes larger than the size of the training volumes (see Table \ref{tab:task01_results_a} and Table \ref{tab:task01_results_b}). Also if the testing size is smaller than the training size, the performance becomes worse (see Table \ref{tab:task01_results_a} and Table \ref{tab:task01_results_c}).  

\begin{table}[!htb]
\caption{P-value between MM and baselines. Non-parametric Mann-Whitney U-Test. 100 labelled slices of CARVE.}
\label{pvalue}
\centering
\begin{tabular}{c  c  c  c  c  c  c  }
    \hline
    \bfseries Sup1 &\bfseries Sup2 & \bfseries MTA &\bfseries MT &\bfseries FM &\bfseries CCT &\bfseries Morph\\
    \hline
     9.13e-5  & 1.55e-2 & 4.5e-3 & 4.3e-4 & 1.05e-2 & 1.8e-3 & 2.2e-3 \\
     \hline
\end{tabular}
\end{table}

\subsection{Ablation Studies}
\begin{figure*}[!t]
\centering
\includegraphics[width=\textwidth]{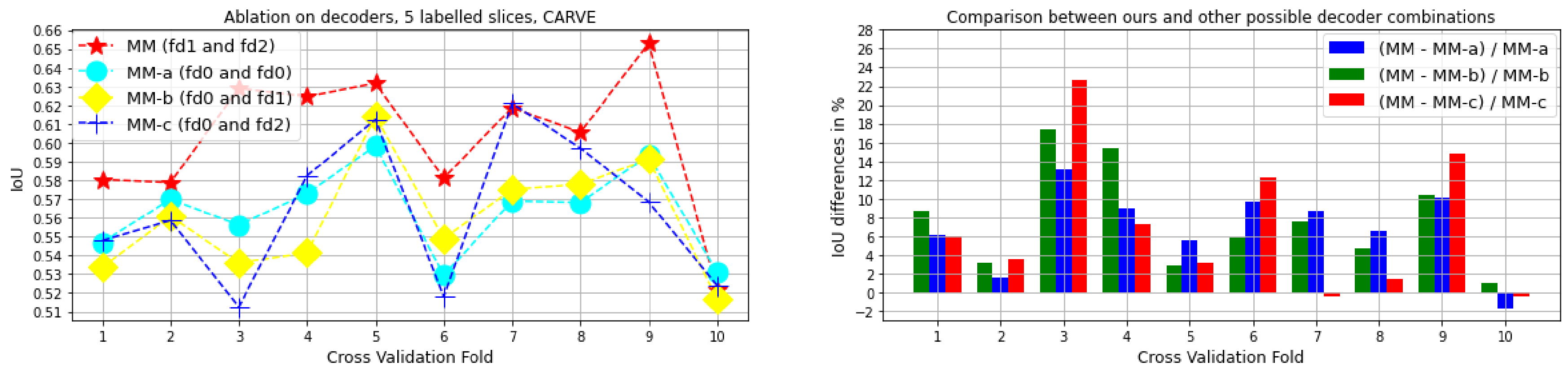}
\caption{Ablation studies on decoder architectures, cross-validation on 5 labelled slices with CARVE. MM is ours.}
\label{fig:ablation_architecture}
\end{figure*}

We performed ablation studies on the architecture of the decoders of MisMatch with cross-validation on 5 labelled slices of CARVE: 1) ``MM-a'', a two-headed U-net with standard convolutional blocks in decoders, the prediction confidences of these two decoders can be seen as both normal confidence, however, they are essentially slightly different because of random initialisation, we denote the decoder of U-net as $f_{d0}$; 2) ``MM-b'', a standard decoder of U-net and a negative attention shifting decoder $f_{d2}$, this one can be seen as between normal confidence and less confidence; 3) ``MM-c'', a standard decoder of U-net and a positive attention shifting decoder $f_{d1}$, this one can be seen as between normal confidence and higher confidence; 4) ``MM'', $f_{d1}$ and $f_{d2}$ (Ours). As shown in Fig \ref{fig:ablation_architecture}, our MisMatch (''MM'') outperforms other combinations in 8 out of 10 experiments and it performs on par with the others in the rest 2 experiments. Among the results when MisMatch outperforms, MisMatch outperforms MM-a by 2\%-14\%; outperforms MM-b by 3\%-18\%; outperforms MM-c by 4\%-22\%. We also tested $\alpha$ at 0, 0.0005, 0.001, 0.002, 0.004 with the same experimental setting. The optimal $\alpha$ appears at 0.002 in Table \ref{tab:alpha}. We also found that gradient cutting helps to improve segmentation performance too, see Table \ref{tab:gradient_cutting}. In terms of network topology, as shown in Table \ref{tab:dilation}
, it seems that larger dilation is not always beneficial.

\begin{table}[!t]
\caption{Ablation studies on alpha value using CARVE with 5 labelled slices.}
\label{tab:alpha}
\centering
\begin{tabular}{c  c  c  c  c  c}
    \hline
    \bfseries alpha  &\bfseries 0.0 &\bfseries 0.0005 & \bfseries 0.001 &\bfseries 0.002 &\bfseries 0.004 \\
    \hline
    \bfseries IoU & 50.75  & 59.16 & 59.45 & 60.25 & 58.89 \\
    \hline
\end{tabular}
\end{table}

\begin{table}[!t]
\caption{Ablation studies on dilation rate in 3D V-net based MisMatch using LA 2018 with 2 labelled cases, $\alpha$ as 1 and cutting gradients, network width 8. Metric is Dice score.}
\label{tab:dilation}
\centering
\begin{tabular}{c  c  c  c  c }
    \hline
    \bfseries Iteration &\bfseries 2000 &\bfseries 3000 & \bfseries 4000 &\bfseries 5000 \\
    \hline
    \bfseries Dilation 6 & 0.6571  & 0.6621& 0.6699 & 0.6561\\
     \bfseries Dilation 9 & 0.7363  & 0.7283 & 0.7180 & 0.6561 \\
     \bfseries Dilation 12 & 0.6980 & 0.6957 & 0.6889 & 0.6561 \\
    \hline
\end{tabular}
\end{table}

\begin{table}[!t]
\caption{Ablation studies on stopping gradients in 3D V-net based MisMatch using LA 2018 with 2 labelled cases, $\alpha$ as 1, network width 8. Metric is Dice score.}
\label{tab:gradient_cutting}
\centering
\begin{tabular}{c  c  c  c  c }
    \hline
    \bfseries Iteration &\bfseries 2000 &\bfseries 3000 & \bfseries 4000 &\bfseries 5000 \\
    \hline
    \bfseries Stop gradient &  0.6896 & 0.7148  & 0.7090 & 0.7057\\
     \bfseries Gradient & 0.6717 & 0.6944 & 0.6952 & 0.6837 \\
    \hline
\end{tabular}
\end{table}

\begin{table}[!t]
\caption{Testing Results on 3D segmenting the whole tumour from Task 01 Brain Tumour from Medical Segmentation Decathlon. Training with learning rate 0.001 and 3500 epochs. Testing on $96 \times 96 \times 96$ cubes. Jac: Jaccard. HD: Hausdorff Distance. ASD: Average Surface Distance.}
\label{tab:task01_results_a}
\centering
\begin{tabular}{c  c c c c}
    \hline
    \bfseries Metrics &\bfseries Dice ($\uparrow$) &\bfseries Jac ($\uparrow$) & \bfseries HD ($\downarrow$) &\bfseries  ASD ($\downarrow$) \\
    \hline
    \bfseries UA-MT & 0.5454 & 0.3864 & 55.25 &  22.74 \\
     \bfseries MisMatch (Ours) & 0.57  & 0.4197 & 49.07 & 22.86 \\
    \hline
\end{tabular}
\end{table}

\begin{table}[!t]
\caption{Testing Results on 3D segmenting the whole tumour from Task 01 Brain Tumour from Medical Segmentation Decathlon. Training with learning rate 0.001 and 3500 epochs. Testing on $48 \times 48 \times 96$ cubes. Jac: Jaccard. HD: Hausdorff Distance. ASD: Average Surface Distance.}
\label{tab:task01_results_b}
\centering
\begin{tabular}{c  c c c c}
    \hline
    \bfseries Metrics &\bfseries Dice ($\uparrow$) &\bfseries Jac ($\uparrow$) & \bfseries HD ($\downarrow$) &\bfseries  ASD ($\downarrow$) \\
    \hline
    \bfseries UA-MT & 0.2926  & 0.1769  & 72.66 & 31.98 \\
    \bfseries MisMatch (Ours) & 0.3133 & 0.1944 & 85.35 & 39.27 \\
    \hline
\end{tabular}
\end{table}

\begin{table}[!t]
\caption{Testing Results on 3D segmenting the whole tumour from Task 01 Brain Tumour from Medical Segmentation Decathlon. Training with learning rate 0.001 and 3500 epochs. Testing on $128 \times 128 \times 96$ cubes. Jac: Jaccard. HD: Hausdorff Distance. ASD: Average Surface Distance.}
\label{tab:task01_results_c}
\centering
\begin{tabular}{c  c c c c}
    \hline
    \bfseries Metrics &\bfseries Dice ($\uparrow$) &\bfseries Jac ($\uparrow$) & \bfseries HD ($\downarrow$) &\bfseries  ASD ($\downarrow$) \\
    \hline
     \bfseries UA-MT  & 0.5945 & 0.4390 & 54.88 & 22.15\\
     \bfseries MisMatch (Ours) & 0.6086  &  0.4650 & 47.66 & 23.58 \\
    \hline
\end{tabular}
\end{table}

\begin{figure*}[!t]
\centering
\includegraphics[width=\linewidth]{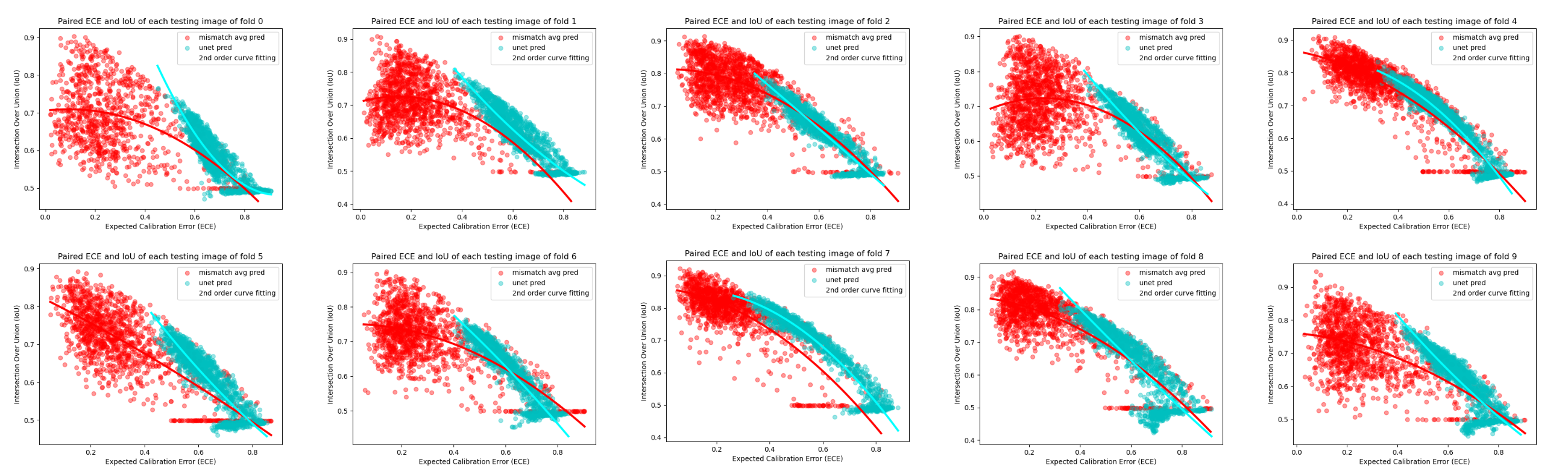}
\caption{Expected calibration error \cite{calibration} against accuracy in 10-fold cross-validation experiments on 50 labelled slices with CARVE. Y-axis: IoU. X-axis: ECE. Each calibration error is calculated from the gap between the confidence and accuracy for each testing image. Each data point in this figure is one testing image. The fitted 2nd order trends of our MisMatch are flatter than U-net, meaning MisMatch is more robust against the calibration error.}
\label{fig:calibration}
\end{figure*}

\begin{figure}[!t]
\centering
\includegraphics[width=0.48\textwidth]{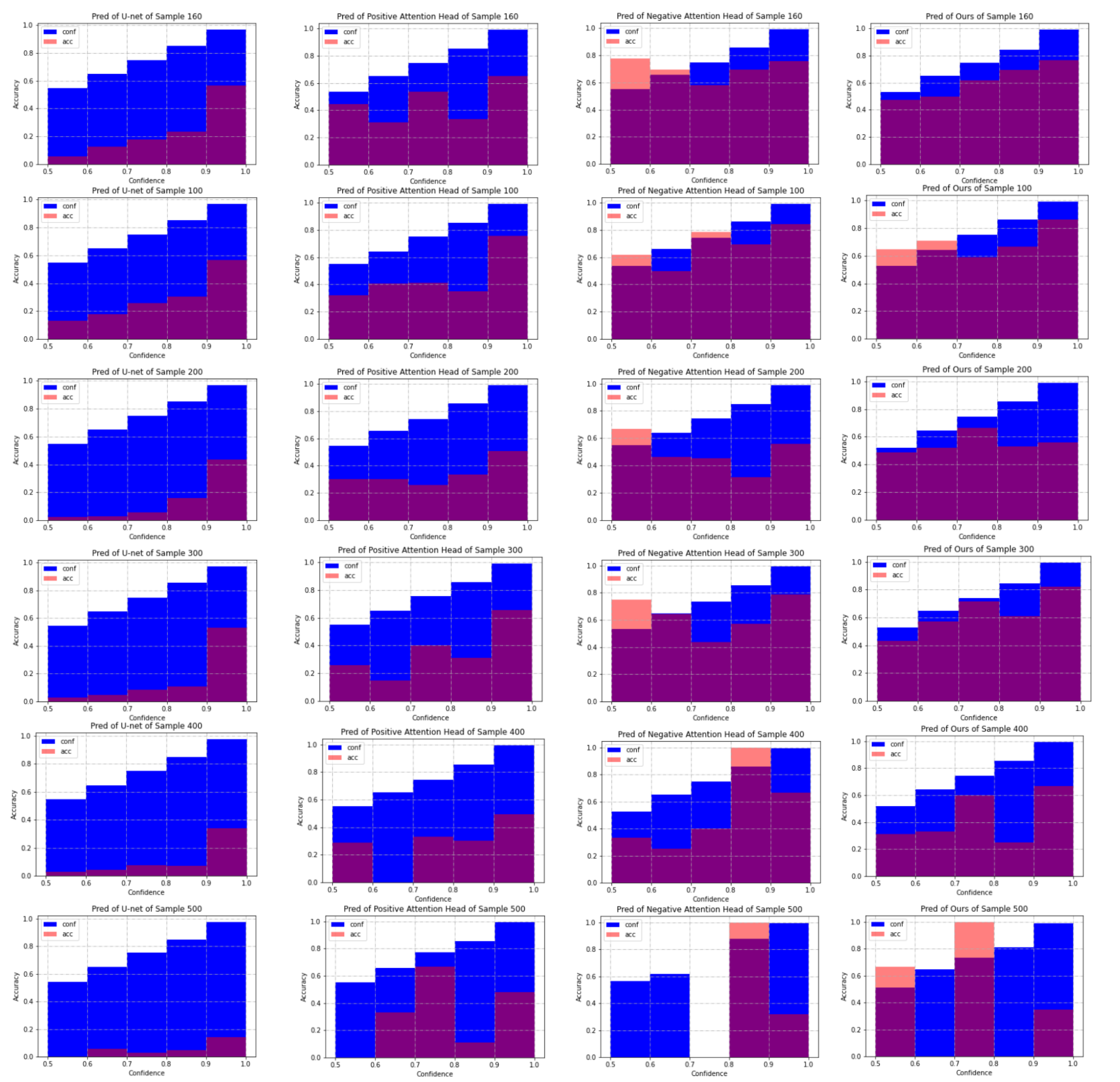}
\caption{Reliability diagrams \cite{calibration} from experiments on 50 labelled slices with CARVE. Blue: Confidence. Red: Accuracy. Each row is on one testing image. X-axis: bins of prediction confidences. Y-axis: accuracy. Column 1: U-net. Column 2: outputs of positive attention decoders. Column 3: outputs of negative attention decoders. Column 4: average outputs of the two decoders. The smaller the gap between the accuracy and the confidence, the better the network is calibrated.}
\label{fig:iou_acc}
\end{figure}

\section{Visualisation of the effectiveness of Learnt Attention Masks}
\label{section:visualisation}
We visualise the confidences of feature maps before and after attention, attention weights and how much the confidences are changed in Fig\ref{fig:attention_visulisation_mismatch} on CARVE. We focus on zoomed-in area of one vessel which is one region-of-interest. As shown in (c) and (e), the confidence outputs between the two decoders are different, the one from the positive attention decoder has more detected high confidence areas on the top of the anatomy of the interest. As illustrated in (j) and (n), the attention weights in the two decoders are drastically different from each other. More specifically, the attention weights in the negative attention decoder have relatively low values around the edges, as shown in green and blue colours, on the contrary, the attention weights in the positive attention decoder have high values in most of the regions of the interest.

Another evidence supporting the effectiveness of attention blocks are the changes of the confidences as shown in (r) and (v). After positive attention weights are applied on (g), it is clear to see in (r) that the surrounding areas of the originally detected contours are now also detected as regions of the interest. Besides, in (v), we observe expected negative changes of the confidences around edges caused by the negative attention shifting. 

The histograms of the feature maps also support the effectiveness of our learnt attention masks. Between the histograms in (j) and (m), for the high confidence interval between 0.9 and 1.0, the negative attention block has more high confidence pixels than the positive attention block. This is because the negative attention block decreases confidence on foreground, thereby ending up with increasing confidence on background, where background class is the majority class naturally containing more pixels than the foreground class.

\section{Confidence and Calibration of Mismatch}
\label{section:calibration}
\textbf{Expected Calibration Error} To qualitatively study the confidence of MisMatch, we adapt two mostly used metrics in the community, which are Reliability Diagrams and Expected Calibration Error (ECE) \cite{calibration}. Following \cite{reliability_diagram}, we first prepare M interval bins of predictions. In our binary setting to classify the foreground, we use 5 intervals between 0.5 to 1. Say $B_{m}$ is the subset of all pixels whose prediction confidence is in interval $I_{m}$. We define accuracy as how many pixels are correctly classified in each interval. The accuracy of $B_{m}$ is formally:
\begin{equation}
\label{acc_ece}
acc(B_{m}) = \frac{1}{|B_{m}|} \sum_{i \in B_{m}} 1 (\hat{y}_{i} = y_{i})
\end{equation} 
Where $\hat{y}_{i}$ is the predicted label and $y_{i}$ is the ground truth label at pixel $i$ in $B_{m}$. The average confidence within $B_{m}$ is defined with the use of $\hat{p}_{i}$ which is the raw probability output of the network at each pixel:
\begin{equation}
\label{conf_ece}
conf(B_{m}) = \frac{1}{|B_{m}|} \sum_{i \in B_{m}} \hat{p}_{i}
\end{equation} 
Ideally, we would like to see $conf(B_{m}) = acc(B_{m})$, which means the network is perfectly calibrated and the predictions are completely trustworthy. To assess how convincing the prediction confidences are, we calculate the gap between confidence and accuracy as Expected Calibration Error (ECE):
\begin{equation}
\label{ece}
ECE = \sum_{m=1}^{M} \frac{|B_{m}|}{n} | acc(B_{m}) - conf(B_{m})|
\end{equation} 

\textbf{MisMatch is well-calibrated and effectively learns to change prediction confidence} As shown in Fig\ref{fig:iou_acc}, both positive attention shifting decoder and negative attention shifting decoder are better calibrated than the plain U-net. Especially, positive attention shifting decoder produces over-confident predictions. Meanwhile, negative attention shifting decoder produces under-confident predictions for a few confidence intervals. This verifies again that MisMatch can effectively learn to differently change the prediction confidences of the same testing images. 

\textbf{Robustness of MisMatch Against Calibration Errors} As shown in the scatter plot (Fig\ref{fig:calibration}) of paired IoU and corresponding Expected Calibration Error (ECE) of all of the testing images in cross-validation experiments on 50 labelled slices of CARVE, higher calibration errors correlate positively with low segmentation accuracy. In general, MisMatch has predictions with less calibration errors and higher IoU values. As shown in the 2nd order regression curves for each trend, MisMatch appears to be more robust against calibration error, as the fitted curve of U-net has a much more steep slope than MisMatch. In other words, with the increase of calibration error, MisMatch suffers less performance drops. 

\section{Limitations and future work}
\textbf{Computational burden} Although MisMatch achieves superior performance over previous methods, it suffers from increased model complexity. Parameter sharing should be incorporated in the future work. For example, the main branch can be shared across the two decoders.

\textbf{Extensions} Future work will extend MisMatch to multi-class 3D tasks. Consistency on multi-class predictions might bring in extra regularisation leading to better performances. We also aim to enhance MisMatch by combining it with existing temporal ensemble techniques \cite{meanteacher}


\section{Conclusion}
We propose MisMatch, an augmentation-free SSL, to overcome the limitations associated with consistency-driven SSL in medical image segmentation. In lung vessel segmentation tasks, the acquisition of labels can be prohibitively time-consuming. For example each case may take 1.5 hours of manual refinement with semi-automatic segmentation\cite{carve2014}. Longer timeframes may be required for cases with severe disease. MisMatch however shows strong clinical utility by reducing the number of training labels requried by more than 90\%. MisMatch requires 100 slices of one case for training whereas the fully labelled dataset comprises 1600 slices across 4 cases. MisMatch when trained on just 10\% of labels achieves a similar performance (IoU: 75\%) to models that are trained with all available labels (IoU: 77\%). 

\printbibliography
\end{document}